\documentclass{llncs}

\usepackage{graphicx}
\usepackage{dsfont}        
\usepackage{exscale}       
\usepackage{amsmath}       
\usepackage{amssymb}
\usepackage{color} 
\usepackage{hyperref} 
\usepackage[all]{hypcap} 
\definecolor{darkred}{rgb}{0.5,0,0}
\definecolor{darkgreen}{rgb}{0,0.5,0}
\definecolor{darkblue}{rgb}{0,0,0.5}

\hypersetup{
linktocpage,
bookmarksnumbered=true,
colorlinks,
linkcolor=darkblue,
filecolor=darkgreen,
urlcolor=darkred,
citecolor=darkblue,
breaklinks=true,
pdfauthor={Olaf Ronneberger (ronneber@informatik.uni-freiburg.de), Philipp Fischer, Thomas Brox},
pdftitle={U-Net: Convolutional Networks for Biomedical Image Segmentation},
pdfsubject={},
raiselinks,
}

\usepackage{booktabs}  

\renewcommand{\vec}[1]{\boldsymbol{\mathbf{#1}}}

\usepackage{makeidx}  
\begin{document}
\frontmatter          
\pagestyle{headings}  

\mainmatter              
\title{U-Net: Convolutional Networks for Biomedical Image Segmentation}
\titlerunning{U-Net}  
%
\author{Olaf Ronneberger \and Philipp Fischer \and Thomas Brox}

%
\authorrunning{O. Ronneberger et al.} 
%
%
\institute{Computer Science Department and BIOSS Centre for Biological Signalling Studies, University of Freiburg, Germany\\
\email{ronneber@informatik.uni-freiburg.de},\\ WWW home page:
\texttt{http://lmb.informatik.uni-freiburg.de/}
}

\maketitle              

\begin{abstract}
There is large consent that successful training of deep networks requires many thousand annotated training samples. In this paper, we present a network and training strategy that relies on the strong use of data augmentation to use the available annotated samples more efficiently. The architecture consists of a contracting path to capture context and a symmetric expanding path that enables precise localization. We show that such a network can be trained end-to-end from very few images and outperforms the prior best method (a sliding-window convolutional network) on the ISBI challenge for segmentation of neuronal structures in electron microscopic stacks. Using the same network trained on transmitted light microscopy images (phase contrast and DIC) we won the ISBI cell tracking challenge 2015 in these categories by a large margin. Moreover, the network is fast. Segmentation of a 512x512 image takes less than a second on a recent GPU. The full implementation (based on Caffe) and the trained networks are available at \href{http://lmb.informatik.uni-freiburg.de/people/ronneber/u-net}{http://lmb.informatik.uni-freiburg.de/people/ronneber/u-net}.
\end{abstract}

\section{Introduction}

In the last two years, deep convolutional networks have outperformed the state of the art in many visual recognition tasks, e.g. \cite{Krizhevsky,R-CNN}. While convolutional networks have already existed for a long time \cite{LeCun_NC1989}, their success was limited due to the size of the available training sets and the size of the considered networks. The breakthrough by Krizhevsky et al.~\cite{Krizhevsky} was due to supervised training of a large network with 8 layers and millions of parameters on the ImageNet dataset with 1 million training images. Since then, even larger and deeper networks have been trained \cite{VGG}.

The typical use of convolutional networks is on classification tasks, where the output to an image is a single class label. However, in many visual tasks, especially in biomedical image processing, the desired output should include localization, i.e., a class label is supposed to be assigned to each pixel. Moreover, thousands of training images are usually beyond reach in biomedical tasks. Hence, Ciresan et al.~\cite{schmidhuber12deepneural} trained a network in a sliding-window setup to predict the class label of each pixel by providing a local region (patch) around that pixel as input. First, this network can localize. Secondly, the training data in terms of patches is much larger than the number of training images. The resulting network won the EM segmentation challenge at ISBI 2012 by a large margin.

Obviously, the strategy in Ciresan et al.~\cite{schmidhuber12deepneural} has two drawbacks. First, it is quite slow because the network must be run separately for each patch, and there is a lot of redundancy due to overlapping patches.
Secondly, there is a trade-off between localization accuracy and the use of context. Larger patches require more max-pooling layers that reduce the localization accuracy, while small patches allow the network to see only little context. More recent approaches \cite{Seyedhosseini2013,hypercolumns} proposed a classifier output that takes into account the features from multiple layers. Good localization and the use of context are possible at the same time.

\begin{figure}[t]
  \centering
  \includegraphics[width=\textwidth]{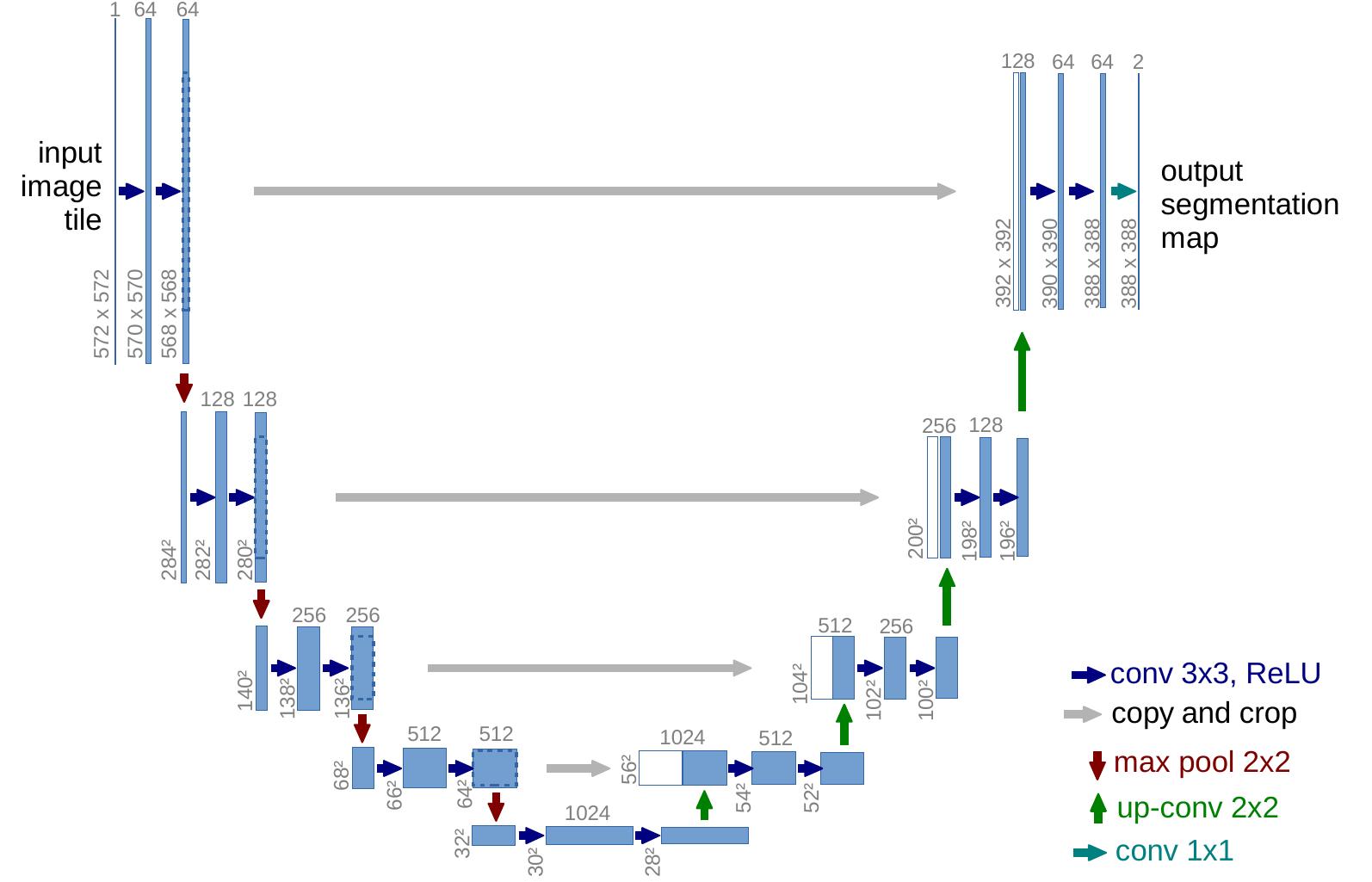}
  \caption{U-net architecture (example for 32x32 pixels in the lowest resolution). Each blue box corresponds to a multi-channel feature map. The number of channels is denoted on top of the box. The x-y-size is provided at the lower left edge of the box. White boxes represent copied feature maps. The arrows denote the different operations. 
  }
  \label{fig:u-net}
\end{figure}

In this paper, we build upon a more elegant architecture, the so-called ``fully convolutional network'' \cite{fullyconv}. We modify and extend this architecture such that it works with very few training images and yields more precise segmentations; see \autoref{fig:u-net}. The main idea in \cite{fullyconv} is to supplement a usual contracting network by successive layers, where pooling operators are replaced by upsampling operators. Hence, these layers increase the resolution of the output. In order to localize, high resolution features from the contracting path are combined with the upsampled output. A successive convolution layer can then learn to assemble a more precise output based on this information.

\begin{figure}[t]
  \centering
  \includegraphics[width=0.7\textwidth]{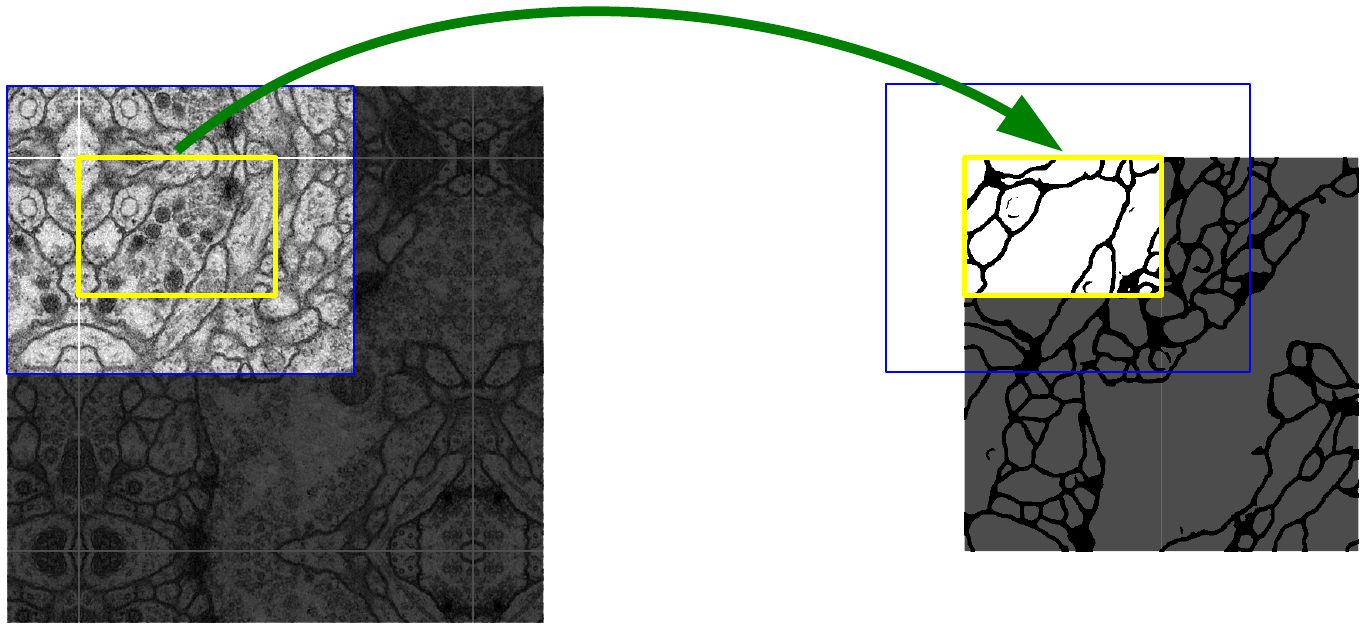}
  \caption{Overlap-tile strategy for seamless segmentation of arbitrary large images (here segmentation of neuronal structures in EM stacks). Prediction of the segmentation in the yellow area, requires image data within the blue area as input. Missing input data is extrapolated by mirroring}
  \label{fig:overlap-tile}
\end{figure}

One important modification in our architecture is that in the upsampling part we have also a large number of feature channels, which allow the network to propagate context information to higher resolution layers. As a consequence, the expansive path is more or less symmetric to the contracting path, and yields a u-shaped architecture. The network does not have any fully connected layers and only uses the valid part of each convolution, i.e., the segmentation map only contains the pixels, for which the full context is available in the input image. This strategy allows the seamless segmentation of arbitrarily large images by an overlap-tile strategy (see \autoref{fig:overlap-tile}). To predict the pixels in the border region of the image, the missing context is extrapolated by mirroring the input image. This tiling strategy is important to apply the network to large images, since otherwise the resolution would be limited by the GPU memory.

As for our tasks there is very little training data available, we use excessive data augmentation by applying elastic deformations to the available training images. This allows the network to learn invariance to such deformations, without the need to see these transformations in the annotated image corpus. This is particularly important in biomedical segmentation, since deformation used to be the most common variation in tissue and realistic deformations can be simulated efficiently. The value of data augmentation for learning invariance has been shown in Dosovitskiy et al.~\cite{unsupervised} in the scope of unsupervised feature learning.

Another challenge in many cell segmentation tasks is the separation of touching objects of the same class; see \autoref{fig:dic-groundtruth}. To this end, we propose the use of a weighted loss, where the separating background labels between touching cells obtain a large weight in the loss function.

The resulting network is applicable to various biomedical segmentation problems. In this paper, we show results on the segmentation of neuronal structures in EM stacks (an ongoing competition started at ISBI 2012), where we outperformed the network of Ciresan et al.~\cite{schmidhuber12deepneural}. Furthermore, we show results for cell segmentation in light microscopy images from the ISBI cell tracking challenge 2015. Here we won with a large margin on the two most challenging 2D transmitted light datasets.

\section{Network Architecture}
The network architecture is illustrated in \autoref{fig:u-net}. It consists of a contracting path (left side) and an expansive path (right side). The contracting path follows the typical architecture of a convolutional network. It consists of the repeated application of two 3x3 convolutions (unpadded convolutions), each followed by a rectified linear unit (ReLU) and a 2x2 max pooling operation with stride 2 for downsampling. At each downsampling step we double the number of feature channels. Every step in the expansive path consists of an upsampling of the feature map followed by a 2x2 convolution (``up-convolution'') that halves the number of feature channels, a concatenation with the correspondingly cropped feature map from the contracting path, and two 3x3 convolutions, each followed by a ReLU. The cropping is necessary due to the loss of border pixels in every convolution. At the final layer a 1x1 convolution is used to map each 64-component feature vector to the desired number of classes. In total the network has 23 convolutional layers.

To allow a seamless tiling of the output segmentation map (see \autoref{fig:overlap-tile}), it is important to select the input tile size such that all 2x2 max-pooling operations are applied to a layer with an even x- and y-size.

\section{Training}
The input images and their corresponding segmentation maps are used to train the network with the stochastic gradient descent implementation of Caffe \cite{Caffe}.
Due to the unpadded convolutions, the output image is smaller than the input by a constant border width. To minimize the overhead and make maximum use of the GPU memory, we favor large input tiles over a large batch size and hence reduce the batch to a single image. Accordingly we use a high momentum (0.99) such that a large number of the previously seen training samples determine the update in the current optimization step.

The energy function is computed by a pixel-wise soft-max over the final feature map combined with the cross entropy loss function. The soft-max  is defined as ${p}_k(\vec{x}) = \exp({a_k(\vec{x})}) / \left(\sum_{k' = 1}^K \exp(a_{k'}(\vec{x}))\right)$
where $a_k(\vec{x})$ denotes the activation in feature channel $k$ at the pixel position $\vec{x} \in \Omega$ with $\Omega \subset \mathbb{Z}^2$. $K$ is the number of classes and ${p}_k(\vec{x})$ is the approximated maximum-function. I.e. ${p}_k(\vec{x}) \approx 1$ for the $k$ that has the maximum activation $a_k(\vec{x})$ and ${p}_k(\vec{x}) \approx 0$ for all other $k$. The cross entropy then penalizes at each position the deviation of ${p}_{\ell(\vec{x})}(\vec{x})$ from 1 using
\begin{equation}
  E = \sum_{\vec{x} \in \Omega} w(\vec{x}) \log({p}_{\ell(\vec{x})}(\vec{x}))
\end{equation}
where $\ell:\Omega \rightarrow \{1,\dots,K\}$ is the true label of each pixel and $w:\Omega \rightarrow \mathds{R}$ is a weight map that we introduced to give some pixels more importance in the training.

\begin{figure}[tbp]
  \centering
  \raisebox{23mm}{\textsf{a }}\includegraphics[height=25mm]{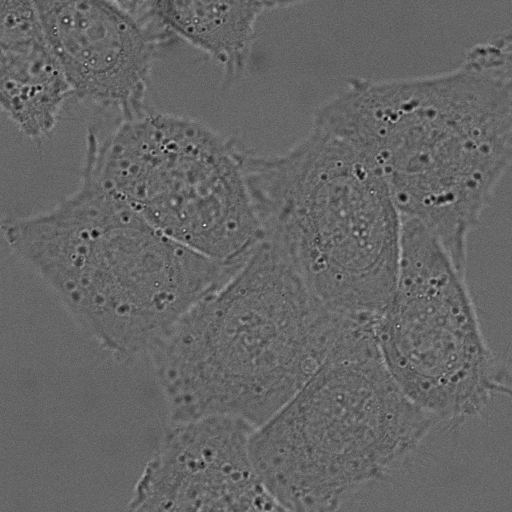}
  \raisebox{23mm}{\textsf{b }}\includegraphics[height=25mm]{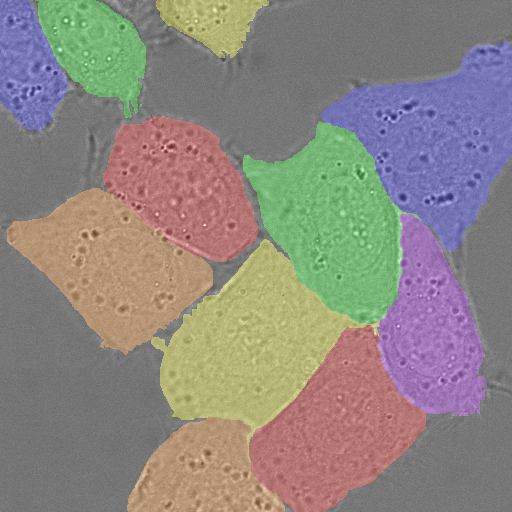}
  \raisebox{23mm}{\textsf{c }}\includegraphics[height=25mm]{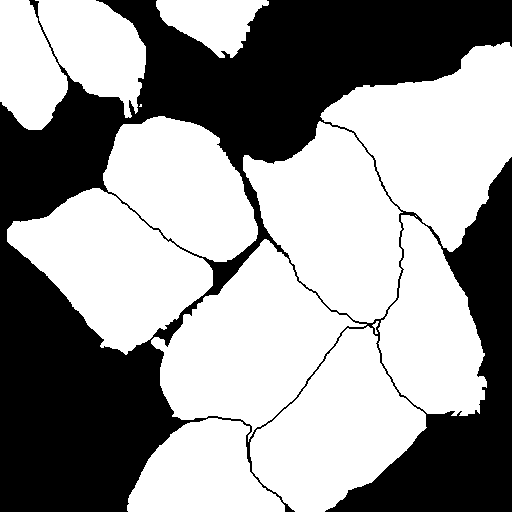}
  \raisebox{23mm}{\textsf{d }}\includegraphics[height=25mm]{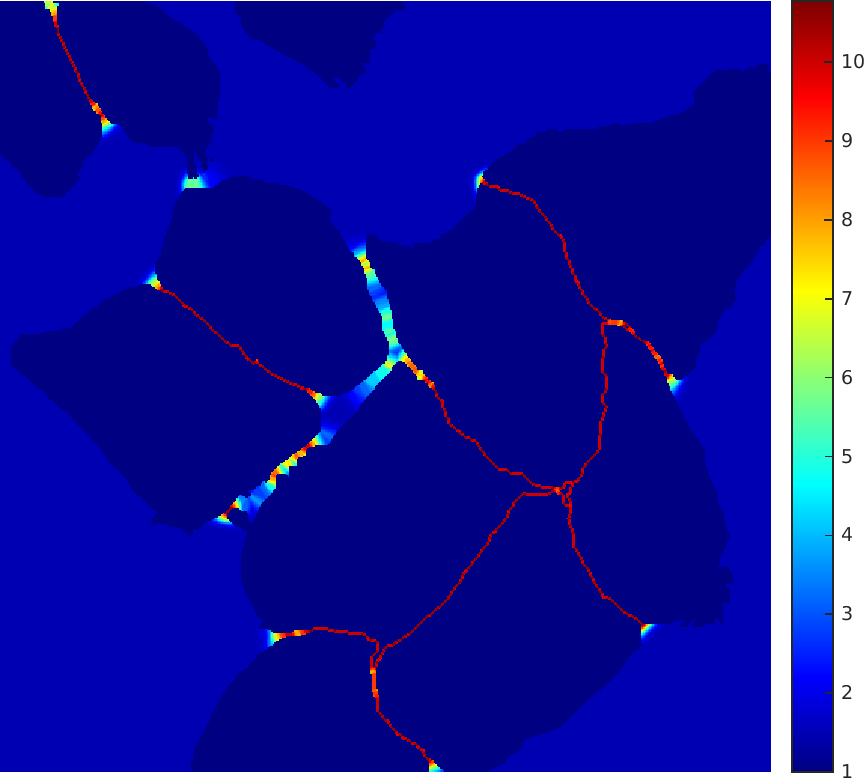}
  \caption{HeLa cells on glass recorded with DIC (differential interference contrast) microscopy. (\textbf{a}) raw image. (\textbf{b}) overlay with ground truth segmentation. Different colors indicate different instances of the HeLa cells. (\textbf{c}) generated segmentation mask (white: foreground, black: background). (\textbf{d}) map with a pixel-wise loss weight to force the network to learn the border pixels.}
  \label{fig:dic-groundtruth}

\end{figure}

We pre-compute the weight map for each ground truth segmentation to compensate the different frequency of pixels from a certain class in the training data set, and to force the network to learn the small separation borders that we introduce between touching cells (See \autoref{fig:dic-groundtruth}c and d).

 The separation border is computed using morphological operations. The weight map is then computed as
\begin{equation}
  w(\vec{x}) = w_c(\vec{x}) + w_0 \cdot \exp\left( - \frac{(d_1(\vec{x}) + d_2(\vec{x}))^2}{2\sigma^2}\right) 
\end{equation}
where  $w_c:\Omega \rightarrow \mathds{R}$ is the weight map to balance the class frequencies, $d_1:\Omega \rightarrow \mathds{R}$ denotes the distance to the border of the nearest cell and $d_2:\Omega \rightarrow \mathds{R}$ the distance to the border of the second nearest cell. In our experiments we set $w_0 = 10$ and $\sigma \approx 5 $ pixels.

In deep networks with many convolutional layers and different paths through the network, a good initialization of the weights is extremely important. Otherwise, parts of the network might give excessive activations, while other parts never contribute. Ideally the initial weights should be adapted such that each feature map in the network has approximately unit variance. For a network with our architecture (alternating convolution and ReLU layers) this can be achieved by drawing the initial weights from a Gaussian distribution with a standard deviation of $\sqrt{2/N}$, where $N$ denotes the number of incoming nodes of one neuron \cite{He2015}. E.g. for a 3x3 convolution and 64 feature channels in the previous layer $N = 9\cdot 64 = 576$.

\subsection{Data Augmentation}
Data augmentation is essential to teach the network the desired invariance and robustness properties, when only few training samples are available. In case of microscopical images we primarily need shift and rotation invariance as well as robustness to deformations and gray value variations. Especially random elastic deformations of the training samples seem to be the key concept to train a segmentation network with very few annotated images.
We generate smooth deformations using random displacement vectors on a coarse 3~by~3 grid. The displacements are sampled from a  Gaussian distribution with 10 pixels standard deviation. Per-pixel displacements are then computed using bicubic interpolation. Drop-out layers at the end of the contracting path perform further implicit data augmentation.

\section{Experiments}

We demonstrate the application of the u-net to three different segmentation tasks. The first task is the segmentation of neuronal structures in electron microscopic recordings. An example of the data set and our obtained segmentation is displayed in \autoref{fig:overlap-tile}. We provide the full result as Supplementary Material. The data set is provided by the EM segmentation challenge~\cite{em-segmentation-webpage} that was started at ISBI 2012 and is still open for new contributions. The training data is a set of 30 images (512x512 pixels) from serial section transmission electron microscopy of the Drosophila first instar larva ventral nerve cord (VNC). Each image comes with a corresponding fully annotated ground truth segmentation map for cells (white) and membranes (black). The test set is publicly available, but its segmentation maps are kept secret. An evaluation can be obtained by sending the predicted membrane probability map to the organizers. The evaluation is done by thresholding the map at 10 different levels and computation of the ``warping error'', the ``Rand error'' and the ``pixel error'' \cite{em-segmentation-webpage}.

The u-net (averaged over 7 rotated versions of the input data) achieves without any further pre- or postprocessing a warping error of 0.0003529 (the new best score, see \autoref{tab:em-results}) and a rand-error of 0.0382. 

\begin{table}[bp]
  \centering
  \caption{Ranking on the EM segmentation challenge \cite{em-segmentation-webpage} (march 6th, 2015), sorted by warping error.}
  \begin{tabular}{r@{~~~}l@{~~~}r@{~~~}r@{~~~}r}
    \toprule
Rank & Group name &	Warping Error &	Rand Error &	Pixel Error \\ 
\midrule
&** human values ** &	0.000005&	0.0021&	0.0010\\
1. & u-net &	\textbf{0.000353}&	0.0382&	0.0611\\
2. & DIVE-SCI& 	0.000355&	0.0305&	0.0584\\
3. & IDSIA \cite{schmidhuber12deepneural}&	0.000420&	0.0504&	0.0613\\
4. & DIVE &	0.000430&	0.0545&	\textbf{0.0582} \\
$\vdots$\\
10.& IDSIA-SCI& 	0.000653&	\textbf{0.0189}	& 0.1027 \\
\bottomrule
  \end{tabular}
  \label{tab:em-results}
\end{table}

This is significantly better than the sliding-window convolutional network result by Ciresan et al.~\cite{schmidhuber12deepneural}, whose best submission had a warping error of 	0.000420 and a rand error of 0.0504.
In terms of rand error the only better performing algorithms on this data set use highly data set specific post-processing methods\footnote{The authors of this algorithm have submitted 78 different solutions to achieve this result.} applied to the probability map of Ciresan et al.~\cite{schmidhuber12deepneural}.

\begin{figure}[tbp]
  \centering
\raisebox{23mm}{\textsf{a}}\includegraphics[height=20mm]{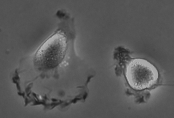}
   \raisebox{23mm}{\textsf{b}}\includegraphics[height=20mm]{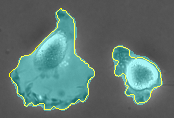}
   \raisebox{23mm}{\textsf{c }}\includegraphics[height=25mm]{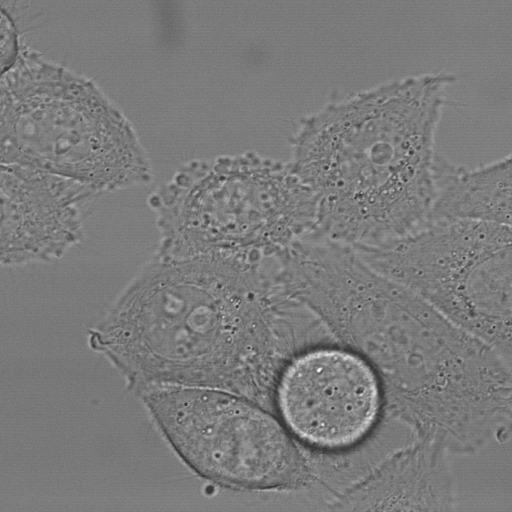}
   \raisebox{23mm}{\textsf{d }}\includegraphics[height=25mm]{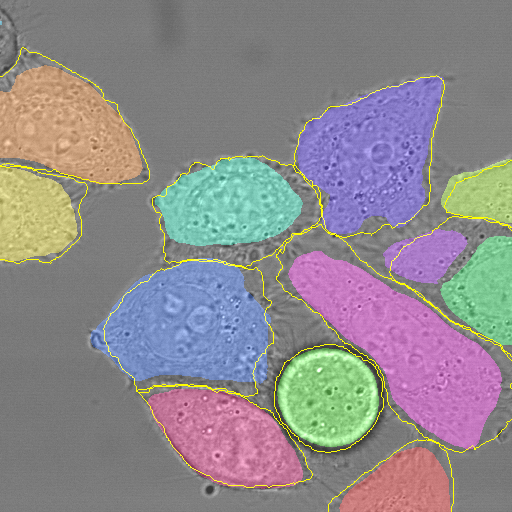}
\caption{Result on the ISBI cell tracking challenge. (\textbf{a}) part of an input image of the ``PhC-U373'' data set. (\textbf{b}) Segmentation result (cyan mask) with manual ground truth (yellow border) (\textbf{c}) input image of the ``DIC-HeLa'' data set. (\textbf{d}) Segmentation result (random colored masks) with manual ground truth (yellow border). }
\label{fig:ph-result}
\end{figure}

We also applied the u-net to a cell segmentation task in light microscopic images. This segmenation task is part of the ISBI cell tracking challenge 2014 and 2015 \cite{Maska2014,cell-tracking-webpage}. The first data set ``PhC-U373''\footnote{Data set provided by Dr. Sanjay Kumar. Department of Bioengineering University of California at Berkeley. Berkeley CA (USA)} contains Glioblastoma-astrocytoma U373 cells on a polyacrylimide substrate recorded by phase contrast microscopy (see \autoref{fig:ph-result}a,b and Supp. Material). It contains 35 partially annotated training images.
Here we achieve an average IOU (``intersection over union'') of 92\%, which is significantly better than the second best algorithm  with 83\% (see \autoref{tab:light-results}).
\begin{table}[tbp]
  \centering
  \caption{Segmentation results (IOU) on the ISBI cell tracking challenge 2015.}
  \begin{tabular}{l@{~~~}l@{~~~}ll}
    \toprule
    Name     &  PhC-U373 & DIC-HeLa \\
    \midrule
    IMCB-SG (2014) & 0.2669    & 0.2935 \\
    KTH-SE  (2014) & 0.7953    & 0.4607 \\
    HOUS-US (2014) & 0.5323    & - \\
    second-best 2015 & 0.83    & 0.46 \\
    u-net (2015)   & \textbf{0.9203}   & \textbf{0.7756} \\
    \bottomrule\\[-0.8cm]
  \end{tabular}
  \label{tab:light-results}
\end{table}
The second data set ``DIC-HeLa''\footnote{Data set provided by Dr. Gert van Cappellen Erasmus Medical Center. Rotterdam. The Netherlands} are HeLa cells on a flat glass  recorded by differential interference contrast (DIC) microscopy (see \autoref{fig:dic-groundtruth}, \autoref{fig:ph-result}c,d and Supp. Material). It contains 20 partially annotated training images. Here we achieve an average IOU of 77.5\% which is significantly better than the second best algorithm with 46\%.

\section{Conclusion}
The u-net architecture achieves very good performance on very different biomedical segmentation applications. Thanks to data augmentation with elastic deformations, it only needs very few annotated images and has a very reasonable training time of only 10 hours on a NVidia Titan GPU (6 GB).
We provide the full Caffe\cite{Caffe}-based implementation and the trained networks\footnote{U-net implementation, trained networks and supplementary material available at \href{http://lmb.informatik.uni-freiburg.de/people/ronneber/u-net}{http://lmb.informatik.uni-freiburg.de/people/ronneber/u-net}}. We are sure that the u-net architecture can be applied easily to many more tasks.

\section*{Acknowlegements}
This study was supported by the Excellence Initiative of
the German Federal and State governments (EXC 294) and by the BMBF (Fkz 0316185B).

\bibliographystyle{splncs03}
\bibliography{Philipps_literature}
\end{document}